\documentclass[10pt,twocolumn,letterpaper]{article}

\usepackage[pagenumbers]{cvpr} 

\usepackage{graphicx}
\usepackage{amsmath}
\usepackage{amssymb}
\usepackage{booktabs}
\usepackage{subcaption}
\usepackage{multirow}
\usepackage{verbatim}
\usepackage{float}
\usepackage{stfloats}

\usepackage[pagebackref,breaklinks,colorlinks]{hyperref}

\usepackage[capitalize]{cleveref}
\crefname{section}{Sec.}{Secs.}
\Crefname{section}{Section}{Sections}
\Crefname{table}{Table}{Tables}
\crefname{table}{Tab.}{Tabs.}

\begin{document}

\title{Delving into Rectifiers in Style-Based Image Translation}

\author{Yipeng Zhang$^{1,2,3}$ \and Bingliang Hu$^{1,2}$ \and Hailong Ning$^{4}$ \and Quang Wang$^{1,2}$~\thanks{Corresponding author, wangquan@opt.ac.cn} \\ 
\and {$^{1}$Xi'an Institute of Optics and Precision Mechanics}
\and {$^{2}$The Key laboratory of Biomedical Spectroscopy of Xi'an}
\and {$^{3}$University of Chinese Academy of Sciences}
\and {$^{4}$Xi’an University of Posts \& Telecommunications}}

\maketitle

\begin{abstract}
	While modern image translation techniques can create photorealistic synthetic images, they have limited style controllability, thus could suffer from translation errors. In this work, we show that the activation function is one of the crucial components in controlling the direction of image synthesis. Specifically, we explicitly demonstrated that the slope parameters of the rectifier could change the data distribution and be used independently to control the direction of translation. To improve the style controllability, two simple but effective techniques are proposed, including Adaptive ReLU (AdaReLU) and structural adaptive function. The AdaReLU can dynamically adjust the slope parameters according to the target style and can be utilized to increase the controllability by combining with Adaptive Instance Normalization (AdaIN). Meanwhile, the structural adaptative function enables rectifiers to manipulate the structure of feature maps more effectively. It is composed of the proposed structural convolution (StruConv), an efficient convolutional module that can choose the area to be activated based on the mean and variance specified by AdaIN. Extensive experiments show that the proposed techniques can greatly increase the network controllability and output diversity in style-based image translation tasks.

\end{abstract}

\section{Introduction}
\label{sec:intro}
\begin{figure}
	\begin{center}
		\includegraphics[width=1.0\linewidth]{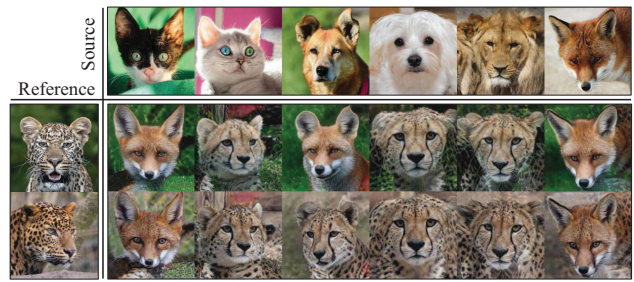}
	\end{center}
	\caption{An example of style inconsistency and translation errors. Source: input content images; Reference: target styles, which are mapped to style codes before sending to the generator. The same reference image, 'leopard,' is translated to both 'leopard' and 'fox,' indicating that the style code has inadequate control over the translation result.}  
	\label{fig:fails}
\end{figure}

The purpose of modern image-to-image translation is to change the visual effect of the source image while preserving the underlying spatial structure. Typically, image translation models attempt to decouple content from style and generate controllable outputs by manipulating the style codes (or attribute codes)~\cite{lee2018diverse, lin2018conditional, li2021image, huang2018multimodal, liu2019few, saito2020coco}. The controllability is crucial for successful image synthesis, as a controllable model enables the style codes to direct the generation, which is valuable in many practical applications~\cite{patashnik2021styleclip, yang2021gan, richardson2021encoding}. The state-of-the-art techniques choose AdaIN~\cite{huang2017arbitrary} or improved techniques~\cite{karras2020analyzing, jing2020dynamic, chandran2021adaptive, lee2021cosmo, liu2021adaattn, xu2021drb} as the style control module, enabling a single generator to produce controllable and diversified synthetic images~\cite{choi2020stargan, karras2019style, zhou2021cocosnet}. However, existing techniques are still constrained by style controllability in certain situations. For instance, the same style code may produce divergent translation results, or the translation may fail entirely, as shown in~\Cref{fig:fails}.

In this work, we examine a component, the rectifiers, a crucial unit in the deep convolutional neural networks, but is usually overlooked in image synthesis tasks. We argue the rectifiers play at least two roles in style-based image translation: (1) Changing the data distribution. For example, when applying Leaky Rectified Linear Unit~(LeakyReLU)~\cite{maas2013rectifier}, the negative part of the feature map is multiplied by a small non-zero number $a$. The mean and variance of the original distribution in the negative part are accordingly altered. (2) Changing the spatial structure directly through the different transformations on the positive and negative parts. For instance, after performing the ReLU activation, only the spatial structure greater than zero is preserved.

This research investigates the influence of rectifiers on both the data distribution and structure in the style-based image translation tasks. For the former, our idea is to enable the synthetic target to determine the slope parameters, allowing the network to select between activation, dormancy, and even reversed activation. For the latter, we introduce a function that can adaptively choose the regions to be activated. The main contributions of this work are summarized as:

(1) We show the influence of activation functions in image synthesis tasks, demonstrating that rectifier parameters can modify the data distribution and independently control the direction of image generation.

(2) We propose an adaptive activation function, AdaReLU, whose slope parameters are dynamically determined by the target style. In addition, AdaReLU can improve the control over style in most cases when combined with AdaIN.

(3) We propose the structural adaptive function, which is composed of the proposed StruConv that can adaptively select the regions to be activated based on the mean and variance specified by AdaIN, thus enables rectifiers to manipulate the structure of feature maps more effectively.

\section{Related works}

\subsection{Style-Based Image Translation}
Early image translation models such as Pix2Pix~\cite{isola2017image}, CycleGAN~\cite{zhu2017unpaired} learned ambiguous mappings from domain $\mathcal{X}$ to domain $\mathcal{Y}$~\cite{zhu2017multimodal}, where an input image $\mathbf{x} \in \mathcal{X}$ had a single and indeterminate counterpart $G(\mathbf{x}) \in \mathcal{Y}$. Zhu~\etal introduced an additional latent code $\mathbf{z} \in \mathcal{Z}$ in BicycleGAN~\cite{zhu2017multimodal}. By learning the relationship between $\mathbf{z}$ and the target domain $\mathcal{Y}$, the generator could produce diverse outputs $G(\mathbf{x},\mathbf{z})$ when given different $\mathbf{z}$. Lee~\etal~\cite{lee2018diverse} divided images into a domain-invariant content space and a domain-specific attribute space, thus allowing manipulation of attributes. Huang~\etal~\cite{huang2017arbitrary} translated the latent code $\mathbf{z}$ into AdaIN parameters, which determined the synthesis directions and improved the controllability of the network. Inspired by Instance Normalization~(IN)~\cite{ulyanov2017improved} and Conditional Instance Normalization~(CIN)~\cite{dumoulin2016a}, AdaIN was initially used in the style transfer~\cite{gatys2016image} algorithm. It is a statistics-based module~\cite{jing2019neural} that first normalizes the channel-wise neurons to a standard normal distribution and then assigns a new mean and variance according to the target style. Mathematically it is given by:

\begin{equation}
	{\operatorname{AdaIN}(\mathbf{x}_{i},\mathbf{y}_{i})=\sigma(\mathbf{y}_{i})\left(\frac{\mathbf{x}_{i} -\mu(\mathbf{x}_{i})}{\sigma(\mathbf{x}_{i})}\right)+\mu(\mathbf{y}_{i})},
\end{equation} where $\mu(\cdot)$ and $\sigma(\cdot)$ denote the mean and variance of the input. Each feature map $\mathbf{x}_{i}$ is normalized separately. In comparison to previous approaches, AdaIN significantly improves the network controllability and the quality of the generated images~\cite{huang2018multimodal, wang2019sdit, choi2020stargan}. Moreover, it has inspired many subsequent image synthesis techniques such as Spatially-Adaptive Normalization~(SPADE)~\cite{park2019semantic}, Semantic Region-Adaptive Normalization~(SEAN)~\cite{zhu2020sean}, Weight Modulation and Demodulation~\cite{karras2020analyzing}, AdaConv~\cite{chandran2021adaptive}, \emph{etc}. 

Typically, the style-based image translation models decompose an image into a content space and a style space or attribute space, then combine different content and style codes to create diverse outputs~\cite{lee2018diverse, huang2018multimodal, lin2018conditional}. The state-of-the-art model SatrGAN v2~\cite{choi2020stargan} is no longer limited to cross-domain translation, it can also translate internal domain images by adjusting the statistics. HiDT~\cite{anokhin2020high} has abandoned domain labels entirely and rely on internal biases to decouple and reconstruct the images. Here, the cross-domain translation means translating the image $\mathbf{x}$ in domain $\mathcal{X}$ to image $G(\mathbf{x)}$ in domain $\mathcal{Y}$, \eg, man to woman, dog to cat. Internal domain translation refers to the translation of images within domain $\mathcal{X}$ or domain $\mathcal{Y}$.  

Our work focuses on further improving the controllability of the style-based image translation network, reducing translation errors, and allowing more semantic changes in the synthesized images.

\subsection{Rectifiers}

\begin{figure}
	\begin{center}
		\includegraphics[width=1.0\linewidth]{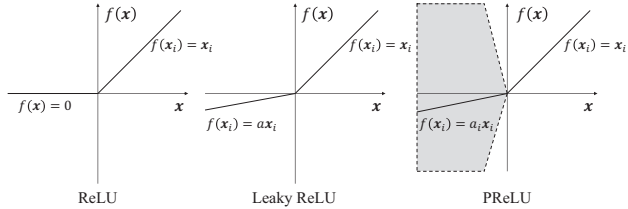}
	\end{center}
	\caption{ReLU, LeakyReLU and PReLU. For LeakyReLU, the slope $a$ is a fixed smaller number for all channels, typically set to $0.2$ in image generation tasks. For PReLU, $a_i$ is learnable and allows each channel $i$ to learn a different slope.}
	\label{fig:relus}
\end{figure}

Rectified Linear Unit (ReLU)~\cite{nair2010rectified} is a nonsaturating activation function which defined as:
\begin{equation}
	{f(\mathbf{x}_{i})=\left\{\begin{array}{cc}
			\mathbf{x}_{i} & \text { if } \mathbf{x}_{i} \geq 0 \\
			0 & \text { if } \mathbf{x}_{i}<0
		\end{array}\right.},
\end{equation} where $\mathbf{x}_{i}$ is the input of the activation $f(\cdot)$ on the $i$th channel.

Employing rectifier~\cite{nair2010rectified, krizhevsky2012imagenet} nonlinearities instead of saturating activation functions (\eg sigmoid, tanh) can avoid gradient exploding/vanishing and accelerate the convergence speed~\cite{krizhevsky2012imagenet}, thus allowing for the construction of very deep networks. However, the gradient in the negative region of the ReLU is zero, which may cause some neurons never being reactivated~\cite{maas2013rectifier}. LeakyReLU~\cite{maas2013rectifier} allows a small gradient for optimization by introducing a small slope parameter, $a$, in the negative region. Mathematically, the LeakyReLU can be formulized as:
\begin{equation}
	{f(\mathbf{x}_{i})=\left\{\begin{array}{cc}
		\mathbf{x}_{i} & \text { if } \mathbf{x}_{i} \geq 0 \\
		a \cdot \mathbf{x}_{i} & \text { if } \mathbf{x}_{i} <0
	\end{array}\right.}.
\end{equation}

PReLU~\cite{he2015delving} introduces a learnable slope parameter $a_i$, improving model fitting and achieving better performance on large-scale classification datasets. PReLU is mathematically similar to LeakyReLU but allows different channels $i$ to learn different parameters.

Generally, a basic modulation block in the statistics-based translator is composed of a convolutional layer, a statistical modification module and a LeakyReLU/ReLU activation function~\cite{liu2019few, saito2020coco, mariotti2021viewnet, mustikovela2021self}. When activated by LeakyReLU, the feature map becomes:

\begin{equation}
	{f(\mathbf{x}_{i,\mu,\sigma})=\left\{\begin{array}{cc}
		\mathbf{x}_{i,\mu,\sigma} & \text { if } x_{i,\mu,\sigma} \geq 0 \\
		\emph{a} \cdot \mathbf{x}_{i,\mu,\sigma} & \text { if } \mathbf{x}_{i,\mu,\sigma}<0
	\end{array}\right.},
\end{equation} where $\mathbf{x}_{i,\mu,\sigma}$ is a feature map specified by any statistics-based modulation module like AdaIN and CIN.

\section{Proposed Method}
\label{sec:method}
In this section, we present the two proposed techniques, AdaReLU and structural adaptation function. AdaReLU is a channel-wise modulation module to control the statistics of neurons; structural adaptation function is a spatial modulation module to select the activated regions based on the statistics. The details are elaborated as follows.

\subsection{Adaptive ReLU}
\label{subsec:adarelu}

\begin{figure}
	\centering
	\begin{subfigure}{0.48\linewidth}
		\begin{center}
			\includegraphics[width=1.0\linewidth]{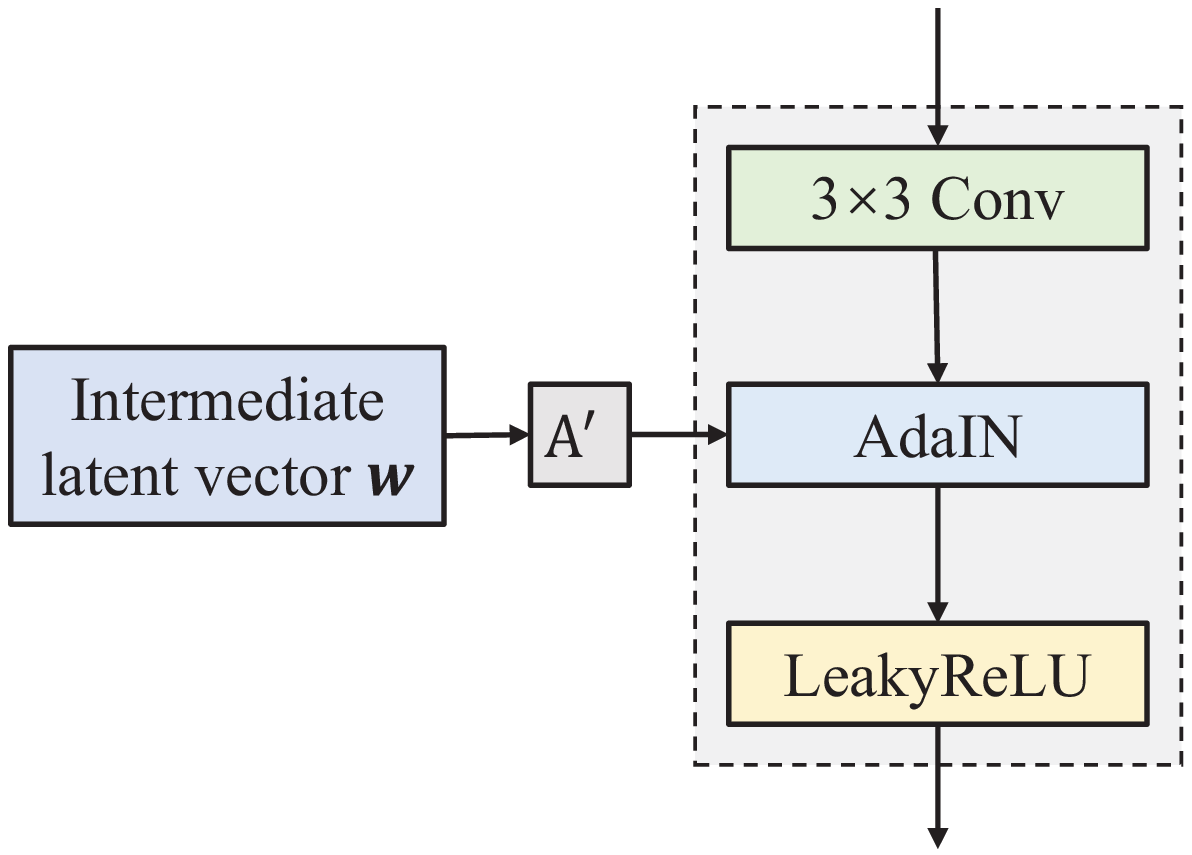}
		\end{center}
		\caption{}
		\label{fig:leakyrelu}
	\end{subfigure}
	\hfill
	\begin{subfigure}{0.48\linewidth}
		\begin{center}
			\includegraphics[width=1.0\linewidth]{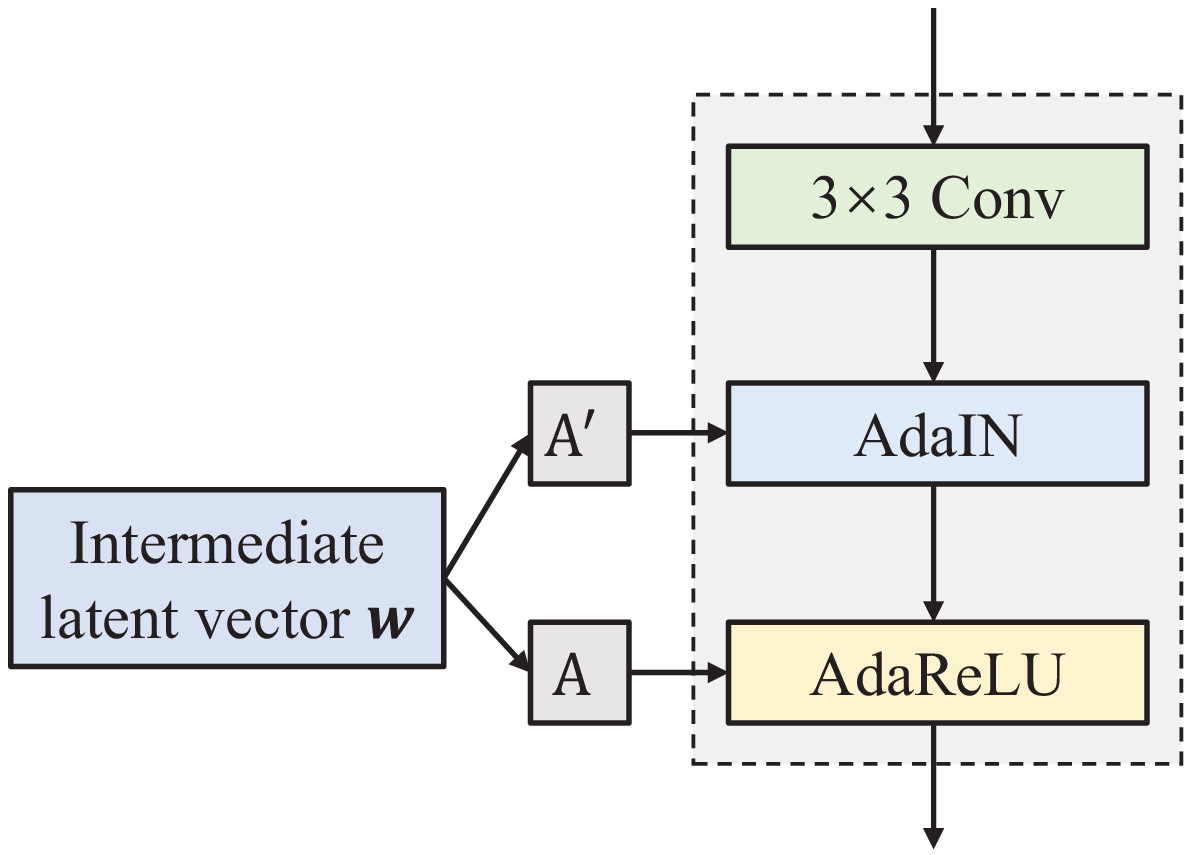}
		\end{center}
		\caption{}
		\label{fig:adarelu}
	\end{subfigure}
	\setlength{\abovecaptionskip}{-1pt}
	\caption{A convolution-normlization-activation block in translator. (a) The most commonly used block in state-of-the-art image generation models. (b) The proposed block which combined AdaIN and AdaReLU.}
	\label{fig2}
	
\end{figure}
The slope parameter of the rectifier is one of the key factors affecting the statistics of the feature map, which is important for a statistic-based style control model. Here, we first propose Adaptive ReLU (AdaReLU). It allows the style code to update the parameter dynamically rather than utilizing fixed numbers. AdaReLU receives an intermediate latent vector $\mathbf{w}$ and converts it to rectifier parameter via an affine transformation, where $\mathbf{w} \in \mathcal{W}$ is mapped from the latent code $\mathbf{z} \in \mathcal{Z}$ or image $\mathbf{x}$ by the mapping network $F$ or style encoder $E$. Mathematically, it is given by:
\begin{equation}  
	{f(\mathbf{x}_i)=\left\{\begin{array}{cc}
		\mathbf{x}_{i} & \text { if } \mathbf{x}_{i} \geq 0 \\
		A(\mathbf{w}) \cdot \mathbf{x}_{i} & \text { if } \mathbf{x}_{i}<0
	\end{array}\right.},
\end{equation} where $A(\cdot)$ denotes an affine transformation.

Intuitively, AdaReLU performs the similar function as AdaIN by scaling magnitude of the feature map, but all of its operations are in the negative part. In~\cref{subsec:adap_relu} we will demonstrate the ability of AdaReLU to manipulate synthesis alone. When AdaIN and AdaReLU are used in combination, the feature map becomes:

\begin{equation}
	{  f(\mathbf{x}_{i,\mu,\sigma})=\left\{\begin{array}{cc}
		\mathbf{x}_{i,\mu,\sigma} & \text { if } \mathbf{x}_{i,\mu,\sigma} \geq 0 \\
		A(\mathbf{w}) \cdot \mathbf{x}_{i,\mu,\sigma} & \text { if } \mathbf{x}_{i,\mu,\sigma}<0
	\end{array}\right.},
\end{equation} where $\mathbf{x}_{i,\mu,\sigma}$ is obtained from the homologous $\mathbf{w}$ via AdaIN:
\begin{equation}
	{\mu, \sigma=A^{\prime}\left(\mathbf{w}\right)} ,
\end{equation}

\begin{equation}
	{\mathbf{x}_{i,\mu,\sigma}=\sigma\left(\frac{\mathbf{x}_i -\mu(\mathbf{x}_i)}{\sigma(\mathbf{x}_i)}\right)+\mu},
\end{equation} where $A^{\prime}$ denotes another learned affine transformation, $\mu$ and $\sigma$ are the target mean and variance.

At this point, a basic convolution-normlization-activation block has additional control over the data distribution. We show the different basic blocks in~\Cref{fig2} and evaluate their style controllability in Section~\ref{sec:results}.

\subsection{Structurual Adaptation Function in Rectifiers}
\label{subsec:adastru}

As previously mentioned, rectifiers can change the structure of the synthesized images directly. Therefore, we aim to explore a function $S$ that enables the network to select the spatial region to be activated based on the input style without changing the data distribution discrepancy before and after performing this function. 

Here, we introduce a specific convolutional module named structural convolution~(StruConv), which modifies the spatial structure of feature maps while maintaining their distribution disparities before and after the operation. StruConv is built on the top of depthwise convolution~(DWConv)~\cite{howard2017mobilenets, chollet2017xception}, which is a part of the depthwise separable convolution~\cite{howard2017mobilenets, chollet2017xception}. A depthwise separable convolution combines a depthwise convolution~(DWConv) and a point-wise convolution~(PWConv) to learn the spatial structure and cross-channel relationship, respectively~\cite{chollet2017xception}. In fact, although DWConv is designed for learning spatial structure only, cross-channel correlation is still affected by it. Assume $c$ is a non-zero constant value and $\mathbf{k}$ is a channel-wise convolution kernel; the $i$\-th feature map, for example, has the same structure after being filtered by $\mathbf{k}_i$ and $c\mathbf{k}_i$, but the data distribution differs. To eliminate the influence of $c$, $\mathbf{k}_i$ is normalized in StruConv:

\begin{equation}
	{\mathbf{k}_{i m n}^{\prime}=\sqrt{\sum_{m,n} \mathbf{k}_{i m n}^{2}}~,} 
\end{equation} where $m$,$n$ represent the height and width of the convolution kernel, respectively.

\begin{figure}[ht]
	\begin{center}
		\includegraphics[width=0.85\linewidth]{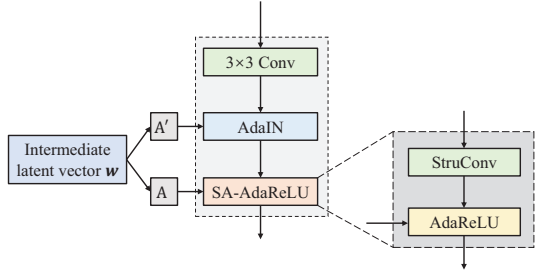}
	\end{center}
	\caption{The proposed architecture with $3\times3$ convolutional layer, AdaIN and SA-AdaReLU.}
	\label{fig:sa-adarelu}
\end{figure}

We use the StruConv as the structural adaptive function $S$. AdaIN is seen as a pre-set condition of StruConv in function $S$, so it is always placed before StruConv.  ~\Cref{fig:sa-adarelu} illustrates a complete Conv$\rightarrow$AdaIN$\rightarrow$SA-AdaReLU block. Mathematically, a Adaptive ReLU with structural adaptation function (SA-AdaReLU) is given by:
\begin{equation}
	{f(\mathbf{x}_{i, \mu,\sigma}, S)=\left\{\begin{array}{cc}
		S(\mathbf{x}_{i,\mu,\sigma}) & \text { if } S(\mathbf{x}_{i,\mu,\sigma}) \geq 0 \\
		A(\mathbf{w}_i) \cdot S(\mathbf{x}_{i,\mu,\sigma}) & \text { if } S(\mathbf{x}_{i,\mu,\sigma})<0
	\end{array}\right.}.
\end{equation} 

Additionally, the structural adaptive functions can be combined with other activation functions. For example, with ReLU, we have the SA-ReLU:
\begin{equation}
	{f(\mathbf{x}_{i, \mu,\sigma}, S)=\left\{\begin{array}{cc}
		S(\mathbf{x}_{i,\mu,\sigma}) & \text { if } S(\mathbf{x}_{i,\mu,\sigma}) \geq 0 \\
		0 & \text { if } S(\mathbf{x}_{i,\mu,\sigma})<0
	\end{array}\right.}.
\end{equation}

\section{Experiment Settings}
\label{sec:experiment_settings}

\begin{figure*}
	\begin{center}
		\includegraphics[width=1.0\linewidth]{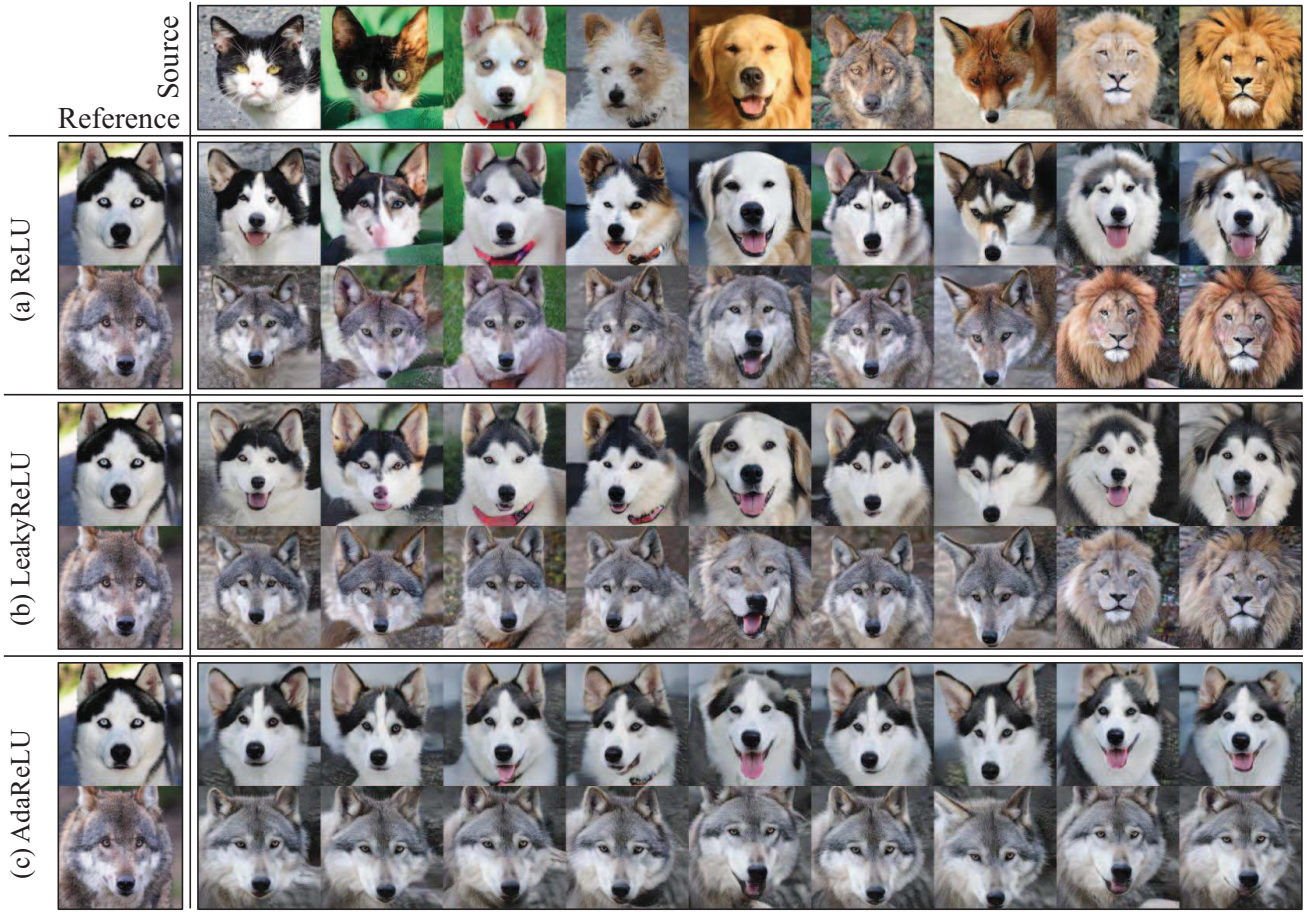}
	\end{center}
	\caption{Qualitative comparison of different activation functions on AFHQ dataset on $N1$.}
	\label{fig:visual_compare}
\end{figure*}

In this section, we describe the baseline activation functions, network structure, dataset, and quantitative evaluation metrics used for the experiments.

\subsection{Baseline activation functions}
 We evaluated the image translation performance under five different activation functions (ReLU, LeakyReLU, PReLU, AdaReLU, SA-AdaReLU) with the same convolutional network structure.  We constructed additional SA-ReLU, SA-LeakyReLU, SA-PReLU in \cref{subsec:sa} to demonstrate the effectiveness of the structural adaptive function $S$. In all experiments, the kernel size of the structural convolution is simply set to $3\times3$.

\subsection{Network Architecture}

We evaluated the effectiveness of our proposed activation function using two generators. The first one, which represented by $N1$, is the state-of-the-art diverse image synthesis model StarGAN v2~\cite{choi2020stargan}, which consists of an encoder without style codes and a decoder with style codes. $N1$ has the residual structure for the AFHQ dataset and skip connections with wing-based heatmap~\cite{wang2019adaptive} for the CelebA-HQ dataset. On all datasets, we use $4$ times downsampling instead of $5$ to avoid excessive information loss of the source image. The second generator is indicated as Network $2$ ($N2$) and consists of an encoder, a decoder, and a translator. Style codes are only used in the translator. The encoder and decoder are constructed using residual blocks, while the translator is constructed from $n$ AdaIN$\rightarrow$Activation$\rightarrow$Conv$3\times3$ blocks. We set $n=6$ in our experiments. The implemented details can be seen in the Appendix. The remainder of the network is the same as  that in~StarGAN v2~\cite{choi2020stargan}, including the style encoder, the mapping network and the discriminator. 

\subsection{Datasets}
The experiments were conducted on the CelebA-HQ\cite{karras2017progressive} and AFHQ\cite{choi2020stargan} datasets. The CelebA-HQ dataset is divided into two categories: men and women, and the AFHQ dataset is divided into three categories: cat, dog, and wild. Each category in AHFQ has several subcategories. In wild, for instance, there are five subcategories: fox, leopard, wolf, tiger, and lion. 

\subsection{Quantitative Evaluation metrics}
We quantitatively evaluated the visual quality, diversity, and controllability of the translated images. The results are quantified for both cross-domain and internal-domain translations. \\

\noindent \textbf{Diversity.} The widely used learned perceptual image patch similarity (LPIPS)~\cite{zhang2018unreasonable} is adopted to quantitatively evaluate the diversity of synthetic images. It was calculated as the $L_1$ distance of the features on the ImageNet pretrained network using ten \emph{different} style codes for the \emph{same} source image. For the latent-guided task, $\mathbf{z}_i$ was randomly sampled by Gaussian distribution. For reference-guided generation, the reference image was randomly sampled. Each source image was translated using $10$ different style codes. \\

\noindent \textbf{Controllability.} When assigning the same style code, the translated semantic information of different source images should be similar. Based on this, we likewise used LPIPS to measure the ability of the style codes to control the translation results. Specifically, we translated $32$ \emph{different} source images using the \emph{same} style code, dividing them into $16$ pairs to compute LPIPS. Each set of $32$ source images was repeated $10$ times with different style codes. \\

\noindent \textbf{Visual Quality.} To quantitatively evaluate the visual quality of the generated images, we calculate the Frechét inception distance (FID)~\cite{heusel2017gans} the same way as StarGAN v2.

\begin{figure*}
	\begin{center}
		\includegraphics[width=1.0\linewidth]{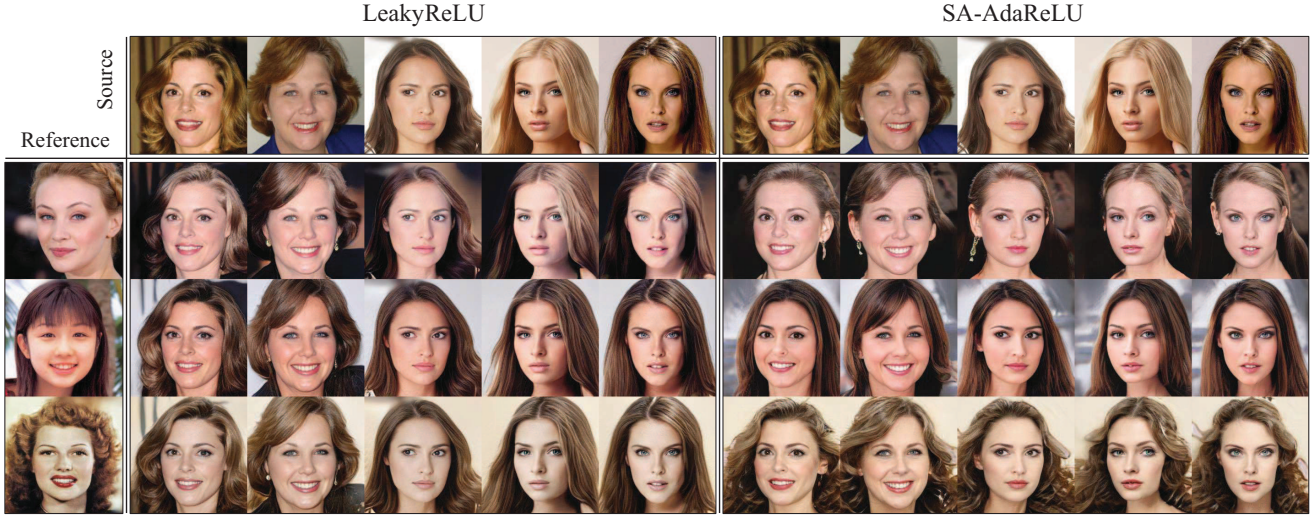}
	\end{center}
	\setlength{\abovecaptionskip}{-1pt}
	\caption{Qualitative comparison of internal domain translation: female to female.}
	\label{fig:visual_celeba}
\end{figure*}

\begin{table*}
	\begin{center}
	\begin{tabular}{|c|c|c|cc|cc|cc|}
		\hline
		\multirow{2}{*}{DataSet}                                                    & \multirow{2}{*}{Domain}                                                          & \multirow{2}{*}{Activation} & \multicolumn{2}{c|}{Visual Quality(FID$\downarrow$)} & \multicolumn{2}{c|}{Diversity(LPIPS$\uparrow$)} & \multicolumn{2}{c|}{Controllability(LPIPS$\downarrow$)} \\
		&                                                                                  &                             & L                         & R                        & L                      & R                      & L                          & R                          \\ \hline
				\multirow{10}{*}{\begin{tabular}[c]{@{}c@{}}CelebA-\\      HQ\end{tabular}} & \multirow{5}{*}{\begin{tabular}[c]{@{}c@{}}Cross-\\      Domain\end{tabular}}    & ReLU                        & 15.3                      & 18.1                     & 0.351                  & 0.322                  & 0.424                      & 0.374                      \\
		&                                                                                  & PReLU                       & 20.5                      & 19.3                     & 0.252                  & 0.287                  & 0.456                      & 0.384                      \\
		&                                                                                  & LeakyReLU                   & 17.3                      & \textbf{17.9}            & 0.263                  & 0.267                  & 0.460                      & 0.387                      \\
		&                                                                                  & AdaReLU                     & \textbf{12.8}             & 18.0                     & 0.412                  & 0.364                  & 0.376                      & 0.345                      \\
		&                                                                                  & SA-AdaReLU                  & 14.2                      & 19.2                     & \textbf{0.440}         & \textbf{0.371}         & \textbf{0.363}             & \textbf{0.320}             \\ \cline{2-9} 
		& \multirow{5}{*}{\begin{tabular}[c]{@{}c@{}}Internal-\\      Domain\end{tabular}} & ReLU                        & 14.2                      & 19.2                     & 0.319                  & 0.294                  & 0.427                      & 0.384                      \\
		&                                                                                  & PReLU                       & 13.3                      & 17.7                     & 0.215                  & 0.246                  & 0.462                      & 0.406                      \\
		&                                                                                  & LeakyReLU                   & 13.1                      & \textbf{16.9}            & 0.214                  & 0.233                  & 0.464                      & 0.409                      \\
		&                                                                                  & AdaReLU                     & \textbf{12.6}             & 19.7                     & 0.396                  & 0.348                  & 0.382                      & 0.361                      \\
		&                                                                                  & SA-AdaReLU                  & 13.7                      & 21.4                     & \textbf{0.428}         & \textbf{0.363}         & \textbf{0.367}             & \textbf{0.331}             \\ \hline
		\multirow{10}{*}{AFHQ}                                                      & \multirow{5}{*}{\begin{tabular}[c]{@{}c@{}}Cross-\\      Domain\end{tabular}}    & ReLU                        & 18.9                      & 19.9                     & 0.426                  & 0.420                  & 0.479                      & 0.450                      \\
		&                                                                                  & PReLU                       & 17.1                      & 19.0                     & 0.437                  & 0.434                  & 0.466                      & 0.441                      \\
		&                                                                                  & LeakyReLU                   & 16.2                      & 19.8                     & 0.448                  & 0.430                  & 0.455                      & 0.444                      \\
		&                                                                                  & AdaReLU                     & 14.9                      & 19.7                     & 0.565                  & 0.530                  & 0.307                      & 0.281                      \\
		&                                                                                  & SA-AdaReLU                  & \textbf{13.9}             & \textbf{18.9}            & \textbf{0.578}         & \textbf{0.544}         & \textbf{0.235}             & \textbf{0.216}             \\ \cline{2-9} 
		& \multirow{5}{*}{\begin{tabular}[c]{@{}c@{}}Internal-\\      Domain\end{tabular}} & ReLU                        & 15.2                      & 16.8                     & 0.388                  & 0.381                  & 0.505                      & 0.476                      \\
		&                                                                                  & PReLU                       & 14.4                      & \textbf{16.5}            & 0.396                  & 0.396                  & 0.491                      & 0.463                      \\
		&                                                                                  & LeakyReLU                   & \textbf{13.1}             & 16.6                     & 0.420                  & 0.399                  & 0.469                      & 0.455                      \\
		&                                                                                  & AdaReLU                     & 14.8                      & 19.9                     & 0.557                  & 0.519                  & 0.313                      & 0.291                      \\
		&                                                                                  & SA-AdaReLU                  & 14.8                      & 19.4                     & \textbf{0.576}         & \textbf{0.543}         & \textbf{0.239}             & \textbf{0.221}             \\ 
		
		\hline

	\end{tabular}
	\caption{Qualitative comparison of different activation functions on $N1$. (L:latent-guided, R:reference-guided)}
	\label{tab:in-across}
\end{center}
\end{table*}

\begin{table*}
	\begin{center}
		\begin{tabular}{|c|c|c|cc|ccc|ccc|}
			\hline
			\multirow{2}{*}{DataSet} & \multirow{2}{*}{Domain}                                                          & \multirow{2}{*}{Activation} & \multicolumn{2}{c|}{Visual   Quality(FID$\downarrow$)} & \multicolumn{3}{c|}{Diversity(LPIPS$\uparrow$)} & \multicolumn{3}{c|}{Controllability(LPIPS$\downarrow$)} \\
			&                                                                                  &                             & w/o SA                     & w/ SA                     & w/o SA            & w/ SA             & Gain    & w/o SA               & w/ SA               & Gain       \\ \hline
						\multirow{8}{*}{\begin{tabular}[c]{@{}c@{}}CelebA-\\      HQ\end{tabular}}  & \multirow{4}{*}{\begin{tabular}[c]{@{}c@{}}Cross-\\      Domain\end{tabular}}    & ReLU     & 17.9                 & 15.8                      & \textbf{0.215}    & \textbf{0.362}    & 0.148   & \textbf{0.260}       & \textbf{0.067}      & 0.193      \\
			&                                                                                  & PReLU                       & \textbf{17.1}              & 15.6                      & \textbf{0.215}    & 0.356             & 0.141   & 0.266                & 0.078               & 0.188      \\
			&                                                                                  & LeakyReLU                   & 17.5                       & 14.2                      & 0.210             & 0.334             & 0.124   & \textbf{0.260}       & 0.119               & 0.141      \\
			&                                                                                  & AdaReLU                     & 17.8                       & \textbf{13.9}             & \textbf{0.215}    & 0.360             & 0.146   & 0.271                & 0.077               & 0.194      \\ \cline{2-11} 
			& \multirow{4}{*}{\begin{tabular}[c]{@{}c@{}}Internal-\\      Domain\end{tabular}} & ReLU                        & 15.4                       & 15.8                      & 0.179             & \textbf{0.362}    & 0.183   & \textbf{0.285}       & \textbf{0.068}      & 0.216      \\
			&                                                                                  & PReLU                       & \textbf{15.0}              & 15.5                      & 0.154             & 0.355             & 0.202   & 0.301                & 0.078               & 0.223      \\
			&                                                                                  & LeakyReLU                   & 15.4                       & \textbf{14.1}             & 0.162             & 0.337             & 0.175   & 0.295                & 0.127               & 0.168      \\
			&                                                                                  & AdaReLU                     & 15.3                       & \textbf{14.1}             & \textbf{0.182}    & 0.352             & 0.175   & 0.286                & 0.081               & 0.209      \\ \hline
			\multirow{8}{*}{AFHQ}    & \multirow{4}{*}{\begin{tabular}[c]{@{}c@{}}Cross-\\      Domain\end{tabular}}    & ReLU    & 17.2                  & 15.6                      & 0.390             & 0.431             & 0.041   & 0.290                & 0.187               & 0.102      \\
			&                                                                                  & PReLU                       & 20.6                       & 15.5                      & 0.356             & 0.402             & 0.045   & 0.301                & 0.271               & 0.030      \\
			&                                                                                  & LeakyReLU                   & 20.1                       & 19.5                      & 0.366             & 0.417             & 0.051   & 0.293                & 0.269               & 0.024      \\
			&                                                                                  & AdaReLU                     & \textbf{14.8}              & \textbf{14.8}             & \textbf{0.416}    & \textbf{0.437}    & 0.021   & \textbf{0.243}       & \textbf{0.152}      & 0.090      \\ \cline{2-11} 
			& \multirow{4}{*}{\begin{tabular}[c]{@{}c@{}}Internal-\\      Domain\end{tabular}} & ReLU                        & 15.1                       & 15.6                      & 0.339             & 0.417             & 0.077   & 0.328                & 0.196               & 0.132      \\
			&                                                                                  & PReLU                       & 19.8                       & \textbf{15.0}             & 0.306             & 0.383             & 0.077   & 0.344                & 0.286               & 0.058      \\
			&                                                                                  & LeakyReLU                   & 17.6                       & 19.1                      & 0.312             & 0.381             & 0.068   & 0.338                & 0.300               & 0.038      \\
			&                                                                                  & AdaReLU                     & \textbf{14.7}              & 15.3                      & \textbf{0.392}    & \textbf{0.431}    & 0.039   & \textbf{0.258}       & \textbf{0.159}               & 0.099      \\ \hline

		\end{tabular}
		\caption{A quantitative comparison with and without the use of structural adaptive functions on $N2$. (Latent-guided, w/o SA: without structural adaptive function; w/ SA: with structural adaptive function.).}
		\label{tab:sa}
	\end{center}
\end{table*}

\section{Results}
\label{sec:results}
In this section, we qualitatively and quantitatively validate the effectiveness of the proposed method. All quantitative results are calculated using the test set, which is unseen during the training.

\subsection{Cross-Domain and Internal-Domain Translation}
\label{subsec:in-cross-trans}

In the AFHQ dataset, subcategory translation such as lion to tiger is classified as internal domain translation. Internal domain translation is more challenging because subcategories are not labeled during the training process. The CelebA-HQ dataset does not have subcategory translation, but the network controllability can be reflected by the semantics change in the internal domain translation.\\

\noindent \textbf{Diversity and Controllability.} We show an example of qualitative comparison results on $N1$ in ~\Cref{fig:visual_compare}. Translation errors exist in the first two activation functions, such as the last two lions in the source images are not being translated successfully. Furthermore, the semantic information may stay unchanged during internal domain translation. When using AdaReLU, all translated results have a consistent color scheme and semantic information. ~\Cref{fig:visual_celeba} is a qualitative comparison of LeakyReLU and SA-AdaReLU for internal domain translation. When using LeakyReLU, the changes are mainly hair color, skin tone, and background color. When SA-AdaReLU is set as the activation function, the hairstyle (short hair$\leftrightarrows$long hair, curly hair$\leftrightarrows$straight hair), overall background, and face are also learned. The quantitative results on $N1$ and $N2$ are shown in~\cref{tab:in-across} and~\cref{tab:sa}, respectively. AdaReLU and SA-AdaReLU improve controllability and output diversity in most cases, with the improvement being more pronounced on the multi-category dataset AFHQ.  \\

\noindent \textbf{Visual Quality.} As shown in~\cref{tab:in-across}, all activation functions can be used to synthesize photorealistic images. The FID of the proposed activations is slightly lower than that of baseline in cross-domain translation but slightly higher than baseline in internal domain translation, probably because the baseline activation methods mainly change only the color theme in this case.

\subsection{Structurual Adaptative Function}
\label{subsec:sa}
In~\cref{tab:sa}, we show the latent-guided quantitative results for the structural adaptive function of the four activation functions (ReLU, LeakyReLU, PReLU, AdaReLU). The experiment was conducted on $N2$. As demonstrated, the structural adaptive function consistently enhances style controllability. While the adaptive slope parameter has little effect on network controllability and output diversity on CelebA-HQ, the structural adaptation increases output diversity by around $100\%$. It is worth noting that in this example, incorporating the structural adaptive function increases the number of parameters by only $0.19\%$ in the translator but significantly improves its performance. 

\subsection{Translation by Adaptive Activation Function}
\label{subsec:adap_relu}

\begin{figure*}
	\begin{center}
		\includegraphics[width=1.0\linewidth]{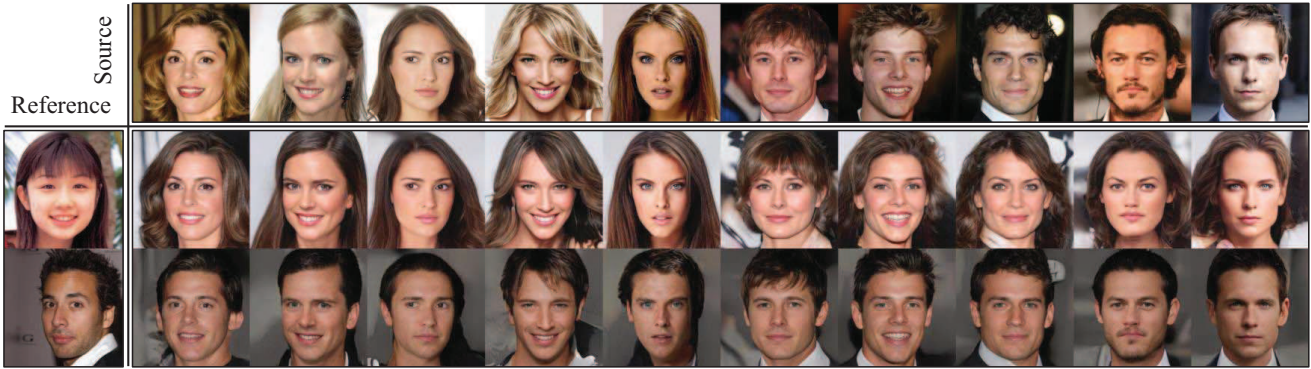}
	\end{center}
	\caption{An example of removing AdaIN and using only AdaReLU as a style control module. AdaReLU can be used independently to direct the translation.}
	\label{fig:celeba_in_adarelu}
\end{figure*}

In~\cref{tab:actv_only} and~\Cref{fig:celeba_in_adarelu}, we perform an experiment on $N1$ that replacing AdaIN by IN and using only the adaptive activation function to control the translation. The difference between using StruConv alone and our proposed structural adaptive function is that the structural adaptive function requires pre-set statistics for the feature maps. As can be seen, despite the decrease in network controllability and output diversity, it can still be used independently to direct the synthetic style. This can explain why the combination of AdaReLU and AdaIN can improve style controllability. It is worth noting that AdaReLU only changes the weight of the channel in the negative part, and its calculation consumption is half of AdaIN. Additionally, AdaReLU with StruConv is always inferior to AdaReLU alone in this experiment, demonstrating the importance of the pre-set mean and variance in structural adaptative function S. At the same time, it reveals the performance improvement of structural adaptive function is not due to the increase in the network complexity.

\begin{table*}
	\begin{center}
		\begin{tabular}{|c|c|c|cc|cc|cc|}
			\hline
			\multirow{2}{*}{DataSet}                                                   & \multirow{2}{*}{Domain}                                                          & \multirow{2}{*}{\begin{tabular}[c]{@{}c@{}}Activation\\      (w/o AdaIN)\end{tabular}} & \multicolumn{2}{c|}{Visual   Quality(FID$\downarrow$)} & \multicolumn{2}{c|}{Diversity(LPIPS$\uparrow$)} & \multicolumn{2}{c|}{Controllability(LPIPS$\downarrow$)} \\
			&                                                                                  &                                                                                        & L                          & R                         & L                      & R                      & L                          & R                          \\ \hline
			\multirow{4}{*}{\begin{tabular}[c]{@{}c@{}}CelebA-\\      HQ\end{tabular}} & \multirow{2}{*}{\begin{tabular}[c]{@{}c@{}}Cross-\\      Domain\end{tabular}}    & AdaReLU                                                                                & \textbf{18.2}              & 18.5                      & \textbf{0.240}         & \textbf{0.236}         & \textbf{0.441}             & \textbf{0.428}             \\
			&                                                                                  & AdaReLU   w/ StruConv                                                                  & 19.3                       & \textbf{18.2}                      & 0.217                  & 0.229                  & 0.445                      & 0.432                      \\ \cline{2-9} 
			& \multirow{2}{*}{\begin{tabular}[c]{@{}c@{}}Internal-\\      Domain\end{tabular}} & AdaReLU                                                                                & \textbf{14.9}              & 16.0                      & \textbf{0.203}         & \textbf{0.197}         & \textbf{0.452}             & \textbf{0.438}             \\
			&                                                                                  & AdaReLU   w/ StruConv                                                                  & 15.3                       & \textbf{15.9}                      & 0.185                  & 0.186                  & 0.455                      & 0.447                      \\ \hline
			\multirow{4}{*}{AFHQ}                                                      & \multirow{2}{*}{\begin{tabular}[c]{@{}c@{}}Cross-\\      Domain\end{tabular}}    & AdaReLU                                                                                & \textbf{30.4}              & \textbf{30.3}             & \textbf{0.372}         & \textbf{0.371}         & \textbf{0.506}             & \textbf{0.486}             \\
			&                                                                                  & AdaReLU   w/ StruConv                                                                  & 47.4                       & 51.4                      & 0.275                  & 0.195                  & 0.537                      & 0.513                      \\ \cline{2-9} 
			& \multirow{2}{*}{\begin{tabular}[c]{@{}c@{}}Internal-\\      Domain\end{tabular}} & AdaReLU                                                                                & \textbf{11.5}              & 11.8                      & \textbf{0.286}         & \textbf{0.286}         & \textbf{0.540}             & \textbf{0.525}             \\
			&                                                                                  & AdaReLU   w/ StruConv                                                                  & \textbf{11.5}              & 12.9                      & 0.168                  & 0.135                  & 0.582                      & 0.553                      \\ \hline
		\end{tabular}
		\caption{Qualitative analysis that removing AdaIN and using only the adaptive activation function as the style control component. AdaReLU w/ StruConv denotes placing StruConv after IN and before AdaReLU.}
		\label{tab:actv_only}
	\end{center}
\end{table*}

\section{Discussion}
\subsection{Replacing StruConv with DWConv}
\begin{figure}
	\begin{center}
		\includegraphics[width=\linewidth]{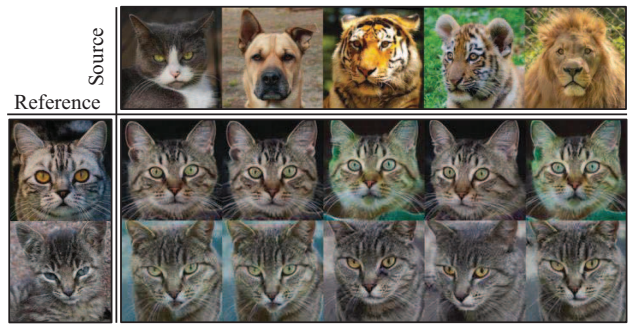}
	\end{center}
	\setlength{\abovecaptionskip}{-0.5pt}
	\caption{Visual examples of using DWConv as the structural adaptive function. It is prone to have color inconsistency in the early stages of training.}
	\label{fig:dwconv}
\end{figure}

When DWConv is used instead of StruConv as the structural adaptation function, it is prone to produce outputs with inconsistent colors in the early stages of training, as shown in~\Cref{fig:dwconv}. Still, these anomalies will decrease with further training. We synthesized 10,000 images using the AFHQ dataset on N1 and manually counted the anomalous images. The percentage of color anomalies was $1.24\%$ for the network using DWConv in the $30K$th iteration and $0.08\%$ in the $100K$th iteration; the network using StruConv was always below $0.1\%$. No significant difference existed in the final model, including the image quality and controllability. This suggests that the network could eliminate the statistical variation introduced by DWConv through an internal learned mechanism.  We think that when the number of iterations and training data are sufficient, the DWConv can be used to replace StruConv.

\subsection{Additional Factors Affect the Style Manipulation}
Many additional factors affect the manipulation of the style, including the objective function, network structure (whether to use residual connections or UNet-like feature fusion), the number of modulation layers and the downsampling times, \emph{etc}.  As shown in~\cref{tab:in-across} and~\cref{tab:sa}, ReLU has insufficient style controllability when used in residual structure but performs well in non-residual-structure networks. We will study the influence of network structure in future works.

\subsection{Avoiding Excessive Style Manipulation}
We observed that, in certain circumstances, the output was practically determined by the input style code when using the proposed activation functions due to their excessive control over the synthesis direction. At this point, the number of down-sampling blocks in the encoder or the modulation layers in the translator can be reduced to weaken the manipulation and therefore balance the source content information and the style information. From this perspective, our suggested activation function could be seen as a means to reduce the computational effort by the same manipulation requiring fewer network layers. We provide several examples in the Appendix.

\section{Conclusion}
\label{sec:conclusion}
In this work, we show that adaptive rectifier parameters and adaptively selected spatial structures can help improve style controllability, reducing translation errors and enhancing semantic manipulation. Accordingly, we propose AdaReLU, which dynamically adjusts the slope parameters according to the target style, and a structural adaptive function that is composed of the efficient StruConv to select the regions to be activated. Extensive experimental results have demonstrated their effectiveness in the style-based image translation tasks. 

\begin{appendix}

\section{Implemented Details} 
For $N1$, we had all the same hyperparameters as in StarGAN v2, and the model was trained for 100K iterations. For N2, the model was trained for 60K iterations with initial weight for diversity sensitive loss~\cite{choi2020stargan}  $\lambda_{ds}=2$ and linearly decayed to $\lambda_{ds}=0.8$ during the training on both CelebA-HQ and AFHQ the datasets. The rest of the hyperparameters were the same as StarGAN v2. All networks were implemented using Pytorch 1.8.2 with Cuda 11.1 on a single RTX 3090 GPU. The system was running Ubuntu 18.04 OS and had an AMD EPYC 7401P CPU @2.0GHz, 64GB of DDR4-2133MHz DRAM. The image resolutions for $N1$ and $N2$ were 256$\times256$ and $128\times128$, respectively. The FID was the minimum value tested on the last five saved models (the model was saved every 2500 iterations), and LPIPS was tested on the final saved model.

\section{Detailed Architecture of Network 2.} 
\begin{figure*}[]
	\begin{center}
		\includegraphics[width=0.95\linewidth]{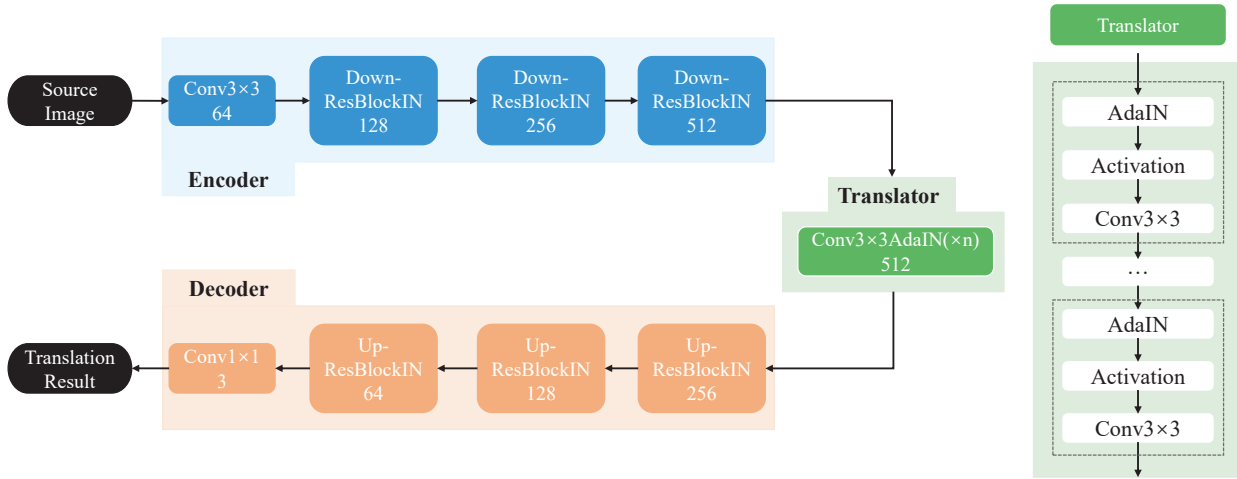}
	\end{center}
	\caption{Generator architecture of network $2$ ($N2$) and details of the translator.}
	\label{fig:n2}
\end{figure*}

\begin{figure}[]
	\begin{center}
		\includegraphics[width=1.0\linewidth]{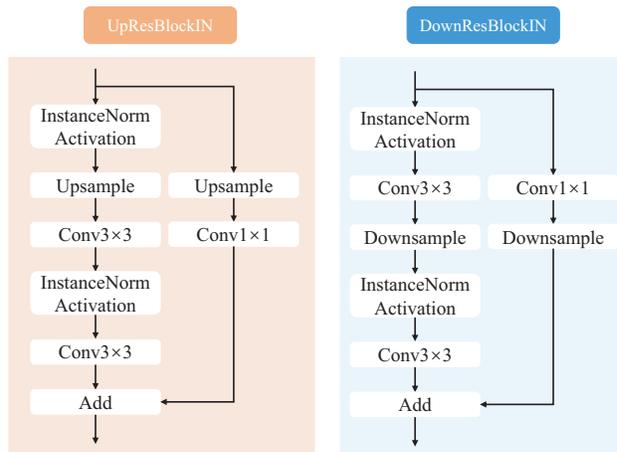}
	\end{center}
	\caption{Detailed architecture of the ResBlocks in $N2$.}
	\label{fig:n2_details}
\end{figure}

\Cref{fig:n2} and~\Cref{fig:n2_details} show the overall architecture and implemented details of Network $2$.

\section{Additional Examples} 

\begin{figure*}[t]
	\begin{center}
		\includegraphics[width=1.0\linewidth]{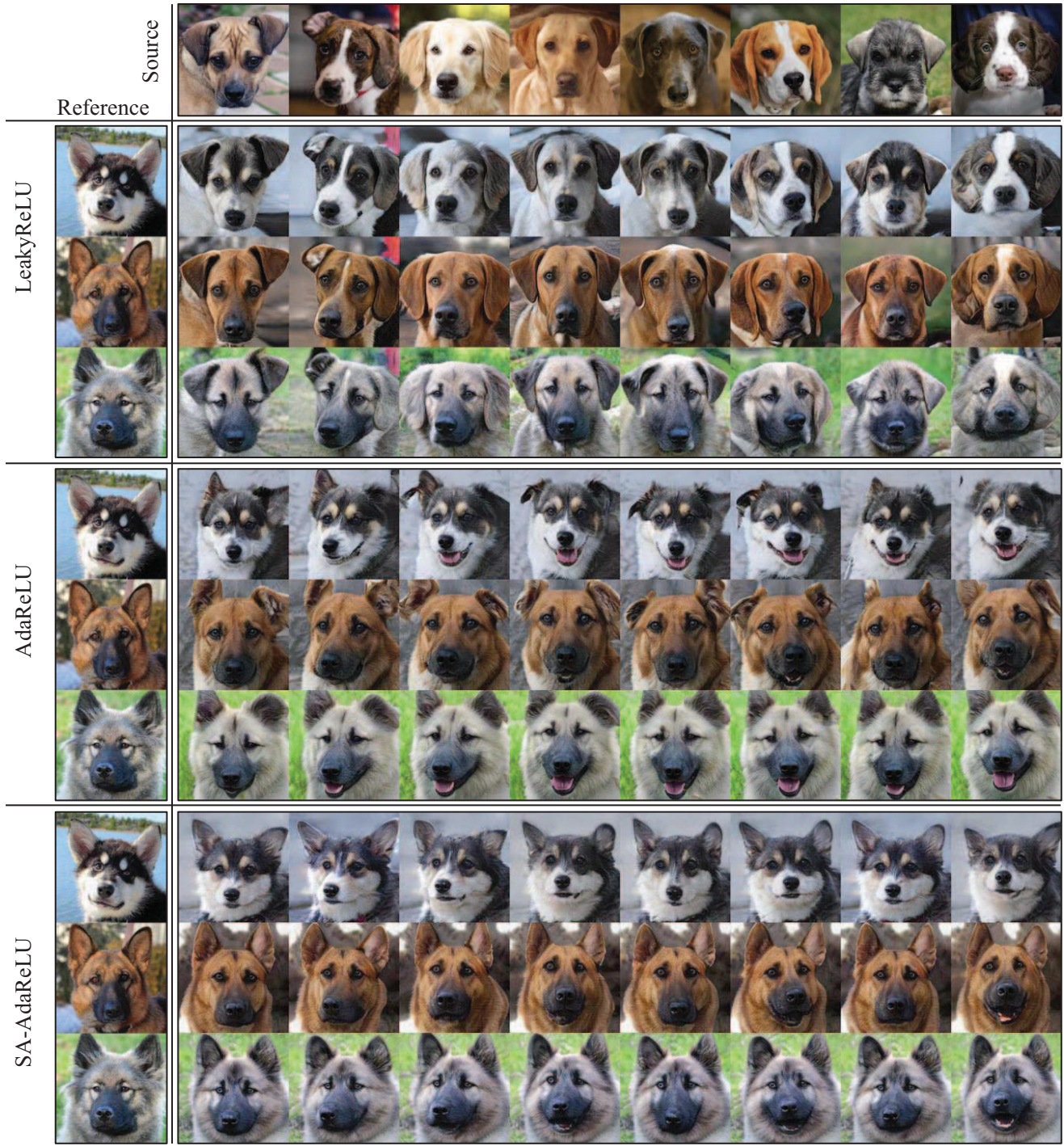}
	\end{center}
	\caption{Dog with 'drooping ears' $\rightarrow$ Dog with 'erect ears' with different rectifiers. ~\\ ~\\ ~\\ ~\\ ~\\ ~\\ ~\\} 
	\label{fig:dog} 
\end{figure*}

\Cref{fig:dog} shows the dog with 'drooping ears' $\rightarrow$ the dog with 'erect ears' on AFHQ dataset, where LeakyReLU is the official released version of StarGAN v2. As can be seen in~\Cref{fig:dog}, the translation is primarily the hair color of the dogs when LeakyReLU is used as the activation function. Although subcategories are translated for both AdaReLU and SA-AdaReLU, when using AdaReLU, the dog’s ears are a hybrid of the two types, resulting in noticeable artifacts. The 'erect ears' are fully translated when SA-AdaReLU is used as the activation function.

\section{The Number of Down-sampling Blocks and Modulation Layers}
\begin{figure*}
	\begin{center}
		\includegraphics[width=\linewidth]{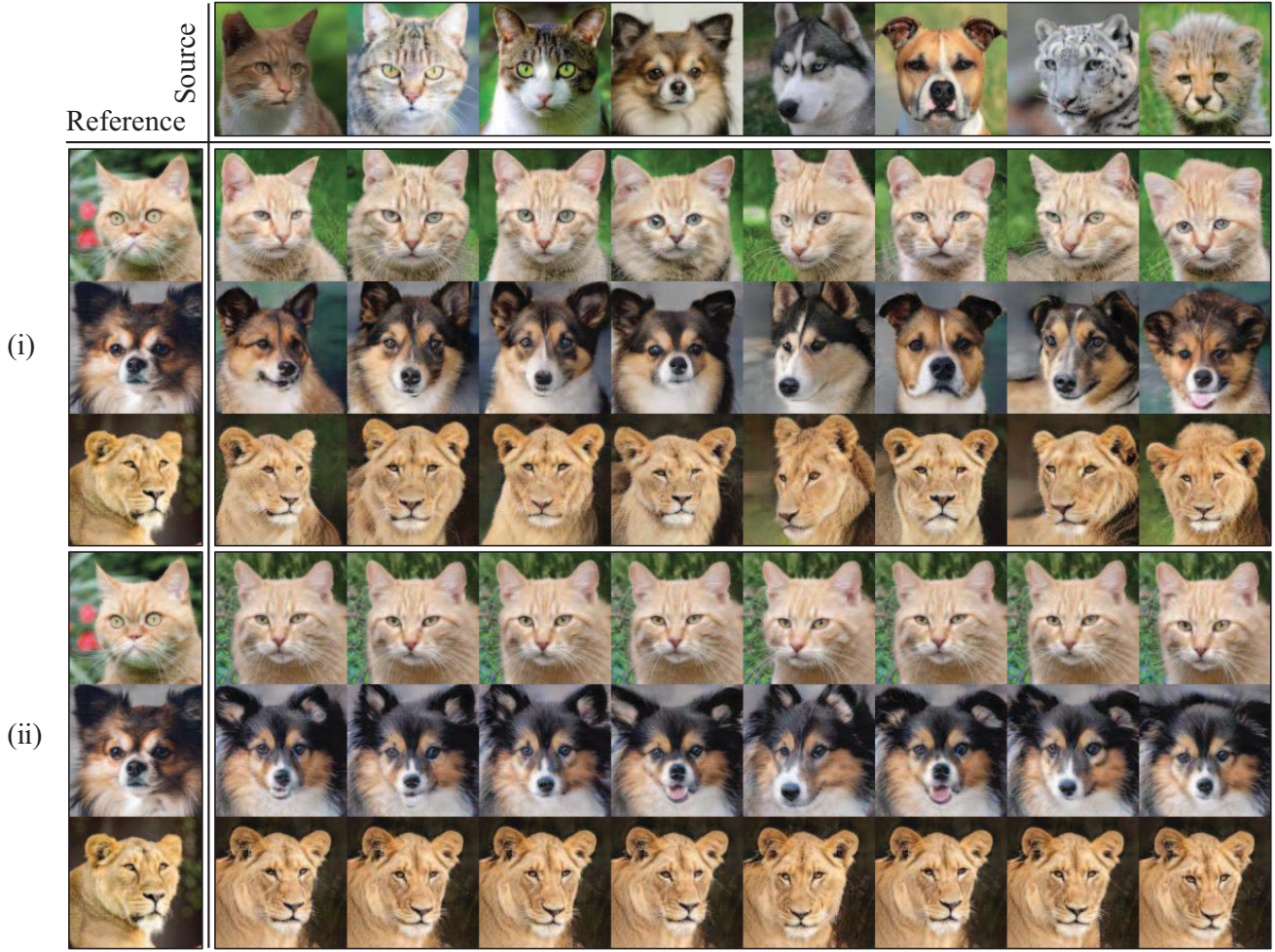}
	\end{center}
	\caption{(\romannumeral1)-(\romannumeral2) The number of down-sampling blocks is 3; The number of modulation layers for~(\romannumeral1) is $4$ and for~(\romannumeral2) is $6$. The results are more controllable with $6$ modulation layers than with $4$ modulation layers, but the outputs are mostly determined by style inputs rather than source images when using $6$ modulation layers. } 
	\label{fig:d3}
\end{figure*}

\begin{figure*}
	\begin{center}
		\includegraphics[width=1.0\linewidth]{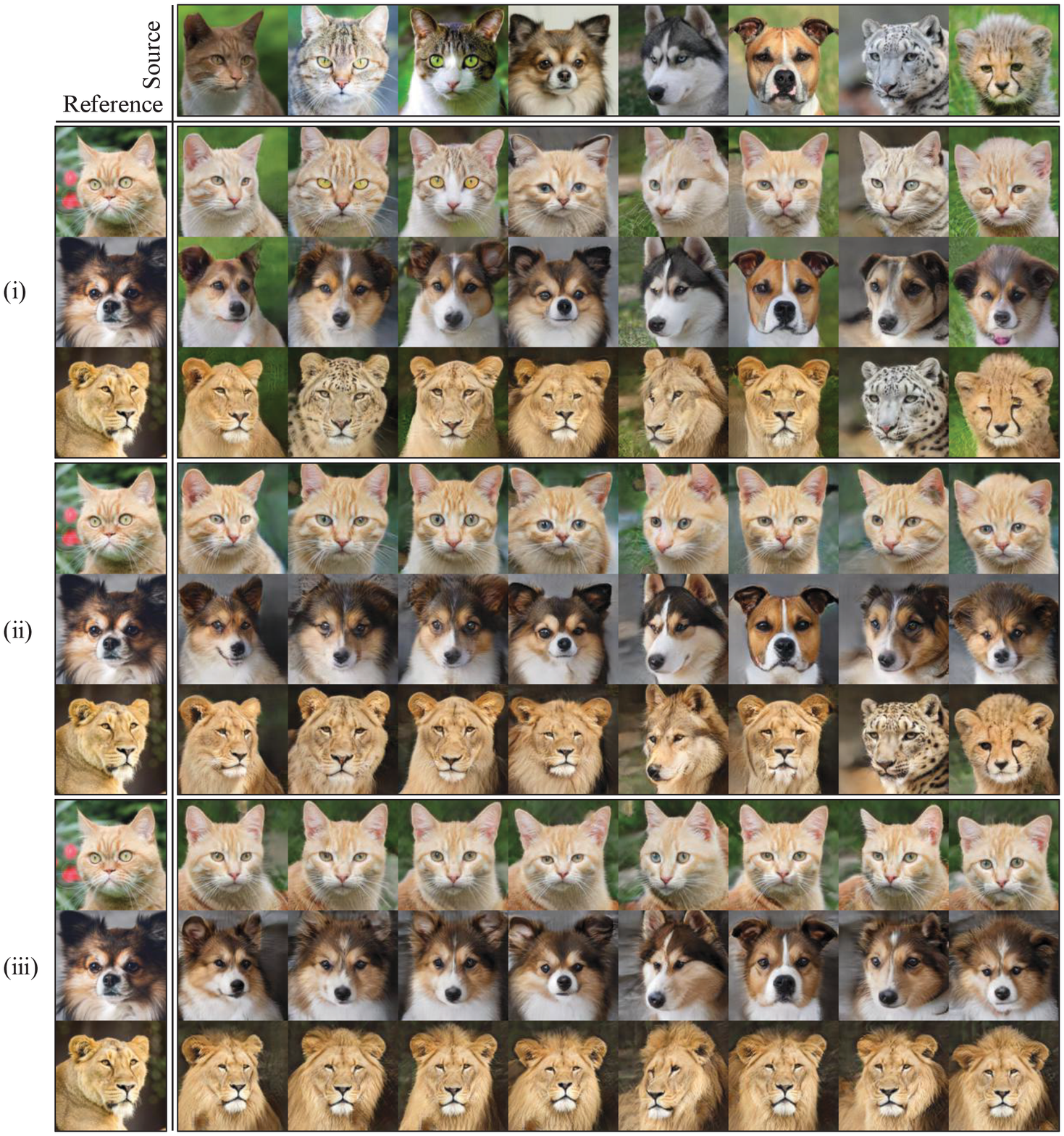}
	\end{center}
	\setlength{\abovecaptionskip}{20pt}
	\caption{(\romannumeral1)-(\romannumeral3) The number of down-sampling blocks is 2. The number of modulation layers for~(\romannumeral1) is $6$ and for~(\romannumeral2) is $8$. The style controllability is insufficient, with obvious translation errors; The number of modulation layers for (\romannumeral3) is 10, with better style controllability. ~\\ ~\\ ~\\ ~\\}
	\label{fig:d2}
\end{figure*}

\Cref{fig:d3},~\Cref{fig:d2} and~\cref{tab:down} are qualitative and quantitative examples performed on $N2$ with SA-AdaReLU activation to show the influence on the number of modulation layers and down-sampling layers.  As can be seen, the more modulation layers present, the more controllable outputs; similarly, the more times of downsampling, the more controllable results. For example, the controllability of a network of ten modulation layers with two times of downsampling is inferior to that of a network of six modulation layers with three times of downsampling.

\begin{table*}[b]
	\centering
	\begin{tabular}{|c|c|c|c|cc|cc|}
		\hline
		\multirow{2}{*}{\begin{tabular}[c]{@{}c@{}}Down-sampling\\      Blocks\end{tabular}} & \multirow{2}{*}{\begin{tabular}[c]{@{}c@{}}Modulation \\      Layers\end{tabular}} & \multirow{2}{*}{Params in G} & \multirow{2}{*}{Params in T} & \multicolumn{2}{c|}{Diversity(LPIPS)} & \multicolumn{2}{c|}{Controllability(LPIPS)} \\
		&                                                                                    &                                  &                                       & L                      & R                      & L                          & R                          \\ \hline
		\multirow{3}{*}{2}                                                                & 6                                                                                  & 14.12M                           & 12.37M                                & 0.179                  & 0.152                  & 0.417                      & 0.400                      \\
		& 8                                                                                  & 19.05M                           & 17.30M                                & 0.300                  & 0.230                  & 0.354                      & 0.355                      \\
		& 10                                                                                 & 23.98M                           & 22.23M                                & 0.385                  & 0.340                  & 0.270                      & 0.255                      \\ \hline
		\multirow{2}{*}{3}                                                                & 4                                                                                  & 17.18M                           & 9.86M                                 & 0.344                  & 0.321                  & 0.322                      & 0.308                      \\
		& 6                                                                                  & 22.11M                           & 14.77M                                & 0.431                  & 0.404                  & 0.159                      & 0.145                      \\ \hline
	\end{tabular}
	\caption{Qualitative internal-domain translation results to show the influence on the number of modulation layers and down-sampling blocks. (G:Generator, T:Translator; L:latent-guided, R:reference-guided) ~\\ ~\\ ~\\ ~\\ ~\\ ~\\}
	\label{tab:down}
\end{table*}

\end{appendix}

{\small
	\bibliographystyle{ieee_fullname}
	\bibliography{ref}

\end{document}